\documentclass[letterpaper, 10 pt, conference]{ieeeconf} 

\usepackage{amsmath,amsfonts}
\usepackage{algorithmic}
\usepackage{algorithm}
\usepackage{array}
\usepackage[]{subfig}
\usepackage{textcomp}
\usepackage{stfloats}
\usepackage{url}
\usepackage{verbatim}
\usepackage{graphicx}
\usepackage{cite}
\usepackage{booktabs}
\usepackage{caption}

\usepackage{color}

\IEEEoverridecommandlockouts                              
\title{\LARGE \bf
OA-ECBVC: A Cooperative Collision-free Encirclement and Capture Approach in Cluttered Environments}
\author{
	Xinyi Wang$^{1*}$, Yulong Ding$^{2*}$, Yizhou Chen$^{1}$, Ruihua Han$^{3}$, Lele Xi$^{4}$, and Ben M. Chen$^{1}$
\thanks{
	The work was supported in part by the Research Grants Council of Hong Kong SAR (14206821 and 14217922), and in part by the Hong Kong Centre for Logistics Robotics, and in part by National Natural Science Foundation of China (Grant No. 62088101).}
\thanks{$^{1}$Xinyi Wang, Yizhou Chen and Ben M. Chen are with the Department of Mechanical and Automation Engineering, The Chinese University of Hong Kong, Shatin, N.T., Hong Kong, China (e-mail: xywangmae@link.cuhk.edu.hk; yzchen@mae.cuhk.edu.hk; bmchen@cuhk.edu.hk);}
\thanks{$^{2}$Yulong Ding is with the Department of Control Science and Engineering, Tongji University, Shanghai, China, and with the Frontiers Science Center for Intelligent Autonomous Systems, Ministry of Education, China; (e-mail: dingyulong@tongji.edu.cn);}
\thanks{$^{3}$Ruihua Han is with Department of Computer Science, The University of Hong Kong, Hong Kong, China; (e-mail: hanrh@connect.hku.hk);}%
\thanks{$^{4}$Lele Xi is with Hebei University of Science and Technology, Shijiazhuang, China; (e-mail: xilele.bit@gmail.com).}%
\thanks{* For equal contributions.}
}

\begin{document}

\maketitle
\thispagestyle{empty}
\pagestyle{empty}

\begin{abstract}
This article investigates the practical scenarios of chasing an adversarial evader in an unbounded environment with cluttered obstacles.  We propose a Voronoi-based decentralized algorithm for multiple pursuers to encircle and capture the evader by reacting to collisions. An efficient approach is presented for constructing an obstacle-aware evader-centered bounded Voronoi cell (OA-ECBVC), which strictly ensures collision avoidance in various obstacle scenarios when pursuing the evader. The evader can be efficiently enclosed in a convex hull given random initial configurations. 
Furthermore, to cooperatively capture the evader, each pursuer continually compresses the boundary of its OA-ECBVC to quickly reduce the movement space of the evader while maintaining encirclement. Our OA-ECBVC algorithm is validated in various simulated environments with different dynamic systems of robots.
Real-time performance of resisting uncertainties shows the superior reliability of our method for deployment on multiple robot platforms.
\end{abstract}

\section{Introduction}

Multi-robot pursuit-evasion (MPE) problem draws considerable research attention in many emergency scenarios, including area surveillance \cite{ref0,semnani2017multi}, target detection and tracking \cite{ref1,khan2022novel}, and wildlife monitoring \cite{ref2}. In such scenarios, the rogue robot may navigate through crowded urban, forests, or other unbounded areas with numerous obstacles, such as buildings, vehicles, pedestrians, and trees. The pursuers must collaborate  effectively to intercept the criminal, prevent their escape, and avoid collisions, which puts high requirements on developing real-time, reactive encirclement strategies that prioritize safety guarantees and efficiency.


 One classic approach to solve MPE problems is to utilize Hamilton-Jacobi Issacs (HJI) partial differential equation (PDE) \cite{ref3,ref4,ref5} based on the formulation of a non-cooperative differential game. However, due to the computational complexity, it encounters huge difficulties in solving HJI equations in the case of multiple players \cite{ref15}. 
Therefore, in order to achieve online and fast pursuit strategies, an increasing number of studies are focusing on using the Voronoi diagram to solve the MPE problems in a decentralized manner.
For example, an area-minimization (AM) pursuit strategy is designed to reduce the movement space of an evader in an obstacle-free environment \cite{ref15,ref16,ref17}. 
 As for more realistic scenarios existing some out-of-range obstacles, a modification of the traditional Voronoi cell method has been proposed, called an obstacle-aware Voronoi cell (OAVC) to avoid collisions when pursuing the evader \cite{pierson2017distributed}. Then, Tian \emph{et al}. \cite{ref20} further considered robot radius into OAVC to expand the safe region for each robot compared with buffered Voronoi cell (BVC) \cite{ref21}.
 However, this method assumes that obstacles are circular in shape, losing generalization in versatile applications and may require additional investigation in the presence of sensor and actuator uncertainties in actual robot systems.
 

Previous works utilizing Voronoi partitions have often been limited by bounded environments and insufficiently enclosed the evader to restrict its movements, resulting in pursuit failures. Due to a lack of encirclement, the evader would be forced into a corner before it can be captured.

Several state-of-the-art studies \cite{ref25,ref26,ref35,ref33} attempted to encircle the adversarial evader within the capture domain by enhancing collaboration among the pursuers.  
 These studies, such as the encirclement-guaranteed partitioning method proposed by Wang \emph{et al.} \cite{ref25} and the distributed surrounding and hunting algorithm by Fang \emph{et al.} \cite{ref26}, relied on angle-based strategies that have strict constraints on the initial spatial positions of the players \cite{ref35} and were vulnerable to various initial configurations.
Furthermore, the lack of consideration for environmental obstacles restricts their practical application.
 In contrast, we construct an evader-oriented Voronoi partition without explicitly constraining angular separation, giving our pursuit strategy greater efficiency and robustness in the face of random initial configurations. 
 
Researchers have recently focused on addressing the challenge of encircling an evader in dynamic and obstacle-rich environments.
  Liao et al. \cite{ref33} designed a simple encirclement approach by evenly spacing the pursuers on a circle at the desired angles. Nevertheless, this deterministic angular separation may lead to lower encirclement and capture efficiency.  Additionally, the obstacle avoidance algorithm they proposed necessitates access to both positional and velocity information via communication, which may not be directly acquirable due to limited onboard resources.	
 

To tackle the challenge of capturing an evader in an unbounded environment with obstacles, this paper proposes a novel pursuit strategy that utilizes the Voronoi partition method to collectively encircle the evader while ensuring collision avoidance, ultimately leading to its capture.
 The main contributions are summarized as follows:
 	\begin{itemize}
   		\item 
An accelerated approach is developed to construct an obstacle-aware evader-centered bounded Voronoi cell (OA-ECBVC). This method combines the separating hyperplane theorem and buffered terms to guarantee collision-free of Voronoi cell  and provide adequate security during pursuit in various obstacle scenarios.
 		\item 
A decentralized approach to encircle and capture an evader is introduced, which allows pursuers to efficiently trap the evader, increasing greater chance of a successful capture in large, open, obstacle-rich areas with random initial configurations.
 		\item 
Simulations are conducted on diverse dynamic systems in complex environments with numerous obstacles to demonstrate the applicability of our method.
Comparisons with other benchmarks highlight its capacity to effectively balance pursuit and collision avoidance. Moreover, our method exhibits superior real-time performance in resisting uncertainties, making it highly reliable for deployment on multiple robot platforms.

 	\end{itemize}
 

The remainder of this paper is organized as follows: Section~\ref{sec: prob} presents the problem formulation, followed by the construction of OA-ECBVC and encirclement and capture strategy in Section~\ref{sec: strategy} and Section~\ref{sec:ec}, respectively. 
Section~\ref{sec:alg} introduces the cooperative pursuit strategy with collision avoidance for different dynamic models. Section~\ref{sec: sim} provides simulation and comparison results and Section~\ref{sec6} conducts the real-time experiments from hardware implementation with mobile robots. Conclusion is given in Section~\ref{sec:con}.

\section{Problem Formulation and Preliminary}
\label{sec: prob}

Consider a MPE problem in an unbounded environment $\mathcal{W} \subseteq  \mathbb{R}^d$, involving a team of $n$ pursuers and one evader and a convex set of obstacles $m$.
$d = 2$ is the dimension of the configuration space. 
Let 
$p_e = (x_{e} ,y_{e} )$ and 
$p_i = (x_{i} ,y_{i} )$
denote the position of the evader and pursuer $i \in \mathcal{I}= \{1,\dots,n \} $, respectively. 
The set of the positions of robots involving the pursuers and the evader is denoted as $\mathcal{P}  =\{p_1,...,p_n, p_e\} \subset \mathcal{W}$.  



\subsection{Pursuit with Collision Avoidance}
Assuming that the evader's policy is unknown, to improve advantages of teamwork, pursuers need to cooperatively form a convex hull to limit the movement space of the evader. 
A convex hull formed by $n_k$ pursuers $z_1,\dots,z_{n_k}$ is denoted as $$
\Omega=\{p \in \mathcal{W}|p =\sum_{i=1}^{n_k}\lambda_ip_{z_i} ,\lambda_i \geq 0, \sum_{i=1}^{n_k}\lambda_i=1\}.    $$
Let $\Vert \cdot \Vert_2$ be the $l_2$ norm. 
The encirclement distance $d_e$ is then defined as  \cite{ref25}:
\begin{equation}
\label{encircledis}
d_{e} :=\left\{
\begin{aligned}
- & \underset{p \in \Omega}{\min} \Vert p -p_e \Vert_{2},& p_e \in \Omega \\
\quad & \underset{p \in \Omega}{\min} \Vert p -p_e \Vert_{2}. & \mbox{otherwise} 
\end{aligned}
\right.
\end{equation}
It is said that the evader is surrounded by the pursuers if either $p_e \in \Omega$ or $d_e \leq 0$ where $t>0$.

Once the encircle the evader, the region where the evader is allowed to freely move may still remain large. As a result, at least one of the pursuers must move close enough to the evader to catch it.
The minimum distance between a team of pursuers and an evader is defined as the capture distance, that is
\begin{equation}
\label{capturedis}
d_{c} :=\underset{i\in\mathcal{I}}{\min}
\Vert p_i -p_e \Vert_{2}.
\end{equation}
It is said that the evader has been captured by the pursuers if the distance between them is smaller than the capture radius $r_c$, which means $d_c \leq r_c$ and $t>0$.
The time at which the encirclement condition and the capture condition hold can be defined as the encirclement time $t_e$ and the capture time $t_c$, respectively. 

Moreover, in the process of pursuing the evader, pursuers are required to avoid collisions with a convex set of obstacles  $\mathcal{O}:=\{O_1,\dots O_m \}$ and other pursuers.
The distance between a pursuer and an obstacle is defined as: \begin{equation}
\label{distobs}
d(p_i,\mathcal{O}_o):=\inf\{\Vert p_i-q_o \Vert | q_o\in \mathcal{O}_o\} .
\end{equation}
where $o\in \mathcal{I}_o=\{1,\dots,m\}$.
The distance between two pursuers is simply given by
\begin{equation}
\label{distpur}
d_{ij} = \Vert p_i -p_j \Vert_{2}.
\end{equation}
 A collision avoidance condition with obstacles and other pursuers can be represented as: 
$d_{io} = d(p_i,\mathcal{O}_o) \textgreater r_i$
and $d_{ij} \textgreater r_i + r_j$, where $r_i$ and $r_j$ are safety radius for pursuer $i$ and $j$, respectively.

\subsection{Problem Formulation}
Given a random initial configuration $\mathcal{P}(0) \in \mathcal{W}$ with $d_{c}(0)> r_c$ find a cooperative trajectory $v_i$ for each pursuer $i$ such that $d_{e}  \leq 0$ and $d_{c}  \leq r_c$ for some $t_c<\infty$ and guarantee $d_{ij} \textgreater r_i + r_j$ and $d_{io} \textgreater r_i$ for any $0 \leq t \leq t_c$ in an unbounded environment.

This challenge involves evader's flexible movements and unbounded environments, making the capture more difficult.
Effective coordination among pursuers is essential for limiting the evader's movements, especially in the case of a more intelligent evader. 
Additionally, MPE games often take place in complex environments, such as forests or urban cities, where generating a collision-free path in the presence of obstacles is crucial. 

\section{Construct ECBVC with Obstacle Awareness}
\label{sec: strategy}

In this section, we present the approach to construct the evader-centered bounded Voronoi cell (ECBVC), followed by introducing the formulation of collision-free Voronoi cell.
\subsection{Evader-centered Bounded Voronoi cell}

Consider the standard Voronoi tessellation, $\mathcal{V}(\mathcal{P}) = \{\mathcal{V}_e,\mathcal{V}_1,...,\mathcal{V}_n\}$ generated by a set of positions of all robots $\mathcal{P}$: 
\begin{equation}
\begin{aligned}
\label{eq:v}
\mathcal{V}_e = & \{p \in \mathcal{W} | \Vert p-p_e\Vert_{2} \leq  \Vert p-p_i\Vert_{2}\},\\
\mathcal{V}_i =  &\{p \in \mathcal{W} | \Vert p-p_i\Vert_{2}\\ 
&\leq  \min\{\Vert p-p_e\Vert_{2},\Vert p-p_j\Vert_{2}\ \}, \forall j\neq i\}
\end{aligned} 
\end{equation}

\begin{figure}[!t]
	\centering
	\setlength{\belowcaptionskip}{-0.5cm}
	\includegraphics[width=2.5in]{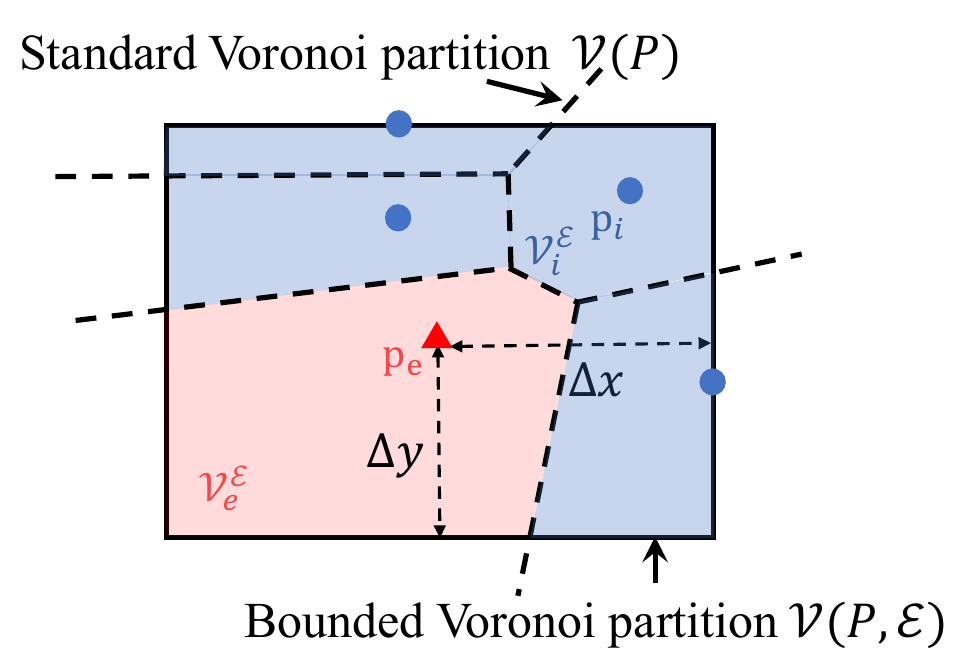}
	\caption{Illustration of ECBVC. The blue circles and red triangles represent the positions of the pursuers and the evader, respectively. The ECBVCs of each pursuer and the evader are shaded in blue and red, respectively.  The bounded Voronoi partition $\mathcal{V}(\mathcal{P},\mathcal{E})$ is evader-oriented, which is determined by the relative position of all players.}
	\captionsetup{justification=centering,margin=0cm}
	\label{fvoronoi2}
\end{figure}

To guide the pursuers to encircle and capture the evader, we introduce the ECBVC for each pursuer which is the intersection of hyperplanes formed by neighboring robots within the limit of a bounded rectangular region. The region denoted as $\mathcal{E}$, is centered at the position of the evader with the lower bound $s_{l}$ and upper bound and $s_{u}$, respectively.
The definition of $\mathcal{E} $ can be given as follows:
\begin{equation}
\begin{aligned}
\label{s}
\mathcal{E}  = \{p  \in \mathcal{W} |\quad & s_{l}  \leq p \leq s_{u} \},
\end{aligned} 
\end{equation}	
where 
\begin{equation}
\begin{aligned}
\label{eq: bound}
s_l  =& [x_e -\Delta x  \quad y_e -\Delta y ]^{\mbox{\footnotesize T}},\\
s_u  =& [x_e +\Delta x  \quad y_e +\Delta y ]^{\mbox{\footnotesize T}}.
\end{aligned}
\end{equation}
The $\Delta x $ and $\Delta y $ represent the half length of the boundaries along the x-axis and y-axis, respectively, as shown in Fig.~\ref{fvoronoi2}.
Let the maximum distance between a team of pursuers and evader as $\Delta x$ and $\Delta y$ along the x-axis and y-axis, respectively: 
\begin{equation}
\begin{aligned}
\label{eq: encircleElength}
& \Delta x  = \underset{i \in \mathcal{I}}{\max}\Vert x_{i}  - x_{e} \Vert_{2},  \\  
& \Delta y  =  \underset{j \in \mathcal{I}}{\max}\Vert y_{j}  - y_{e} \Vert_{2}.
\end{aligned} 
\end{equation}
Thus, given $\mathcal{P}$ and $\mathcal{E}$,
the evader-centered bounded Voronoi tessellation $\mathcal{V}(\mathcal{P},\mathcal{E}) =  \{\mathcal{V}_e^\mathcal{E},\mathcal{V}_1^\mathcal{E},...,\mathcal{V}_n^\mathcal{E}\}$ can be defined as :
\begin{equation}
\begin{aligned}
\label{eq:v2}
\mathcal{V}_e^{\mathcal{E}} =  & \{p \in \mathcal{E} | \Vert p-p_e\Vert_{2} \leq  \Vert p-p_i\Vert_{2}\},\\
\mathcal{V}_i^{\mathcal{E}} =  & \{p \in \mathcal{E} | \Vert p-p_i\Vert_{2}\\
& \leq  \min\{\Vert p-p_e\Vert_{2},\Vert p-p_j\Vert_{2}\ \}, \forall j\neq i\}.
\end{aligned} 
\end{equation}

\subsection{Collision-free Pursuit Region}
To ensure safety during pursuit tasks, it is necessary to divide the pursuer's dominance area into secure zones. This can be accomplished by modifying  $\mathcal{V}(\mathcal{P},\mathcal{E})$ into an obstacle-free region at discrete time intervals $\Delta t$. Strictly collision avoidance between pursuers or obstacles can be achieved if each pursuer remains within a secure Voronoi cell throughout $[t_0, t_0+\Delta t]$.

For each time $t$, the Voronoi cell for each pursuer is determined only by neighboring pursuers and the obstacles, and thus can be formed as the intersection of the following half-spaces \cite{ref27}:
1) $n$ half-spaces that separate robot $\kappa_1$ from robot $\kappa_2$ with parameters of separating hyperplanes $a_{\kappa_1 \kappa_2}$ and $b_{\kappa_1 \kappa_2}$, where $\kappa_1,\kappa_2 \in \{i,j,e\}$ and $\kappa_1 \neq \kappa_2$;
 2) $m$ half-spaces that separate pursuer $i$ from obstacles $\mathcal{O}$ with parameters of separating hyperplanes $a_{io}$ and $b_{io}$, where $o \in \mathcal{I}_o$.

For internal collision avoidance, we can calculate $a_{\kappa_1 \kappa_2}$ and $b_{\kappa_1 \kappa_2}$ by finding the perpendicular line between any two positions of robot $p_{\kappa_1}$ and $p_{\kappa_2}$:
\begin{equation}
 \begin{aligned}
 \label{ai}
&a_{\kappa_1 \kappa_2} = p_{\kappa_1 \kappa_2}=p_{\kappa_1 }-p_{\kappa_2}\text{,}\\
&b_{\kappa_1 \kappa_2} = p_{\kappa_1 \kappa_2}^{\mbox{\footnotesize T}}\frac{p_{\kappa_1}+p_{\kappa_2}}{2}\text{.}
\end{aligned}
\end{equation}
To avoid collisions with obstacles, we consider the obstacle as a bounded convex hull determined by a vertex vector $\Psi_o = [\psi_1, \dots, \psi_{n_o}] \in \mathbb{R}^{d \times n_o}$. The parameter $a_{io}$ can be calculated efficiently by solving the following low-dimensional quadratic programming (QP) problem:
\begin{equation}
 \begin{aligned}
 \label{ao}
 \min \quad &a_{io}^{\mbox{\footnotesize T}}a_{io}\\
 \text{s.t.} \quad & (\psi_l-p_i)^{\mbox{\footnotesize T}}a_{io} \geq 1, \quad \forall l \in \{1,\dots,n_o\}
\end{aligned}
\end{equation}
We then shift  the hyperplane to be tight with the obstacle. Thus, $b_{io} =\min{a_{io}^{\mbox{\footnotesize T}}\psi_o}$.


Furthermore, we employ the idea of safety buffered term \cite{ref21} into original ECBVC. The buffered ECBVC can be interpreted as retracting the edges of its corresponding ECBVC $\mathcal{V}_i^{\mathcal{E}}$ by a safety distance considering the geometric size of the pursuer. 
Denote the following modified buffer term $\beta_{ij} =  r_i \Vert p_i-p_j \Vert_{2}$ and $\beta_{io} =  r_i$ to divide each pursuer into ensure the whole physical body of each pursuer within its corresponding Voronoi cell. 

Based on the above, the definition of evader-centered bounded Voronoi tessellation with obstacle awareness  $\mathcal{V}(\mathcal{P},\mathcal{E},\mathcal{O}) =  \{\mathcal{V}_e^{\mathcal{E},b},\mathcal{V}_1^{\mathcal{E},b},...,\mathcal{V}_n^{\mathcal{E},b}\}$ can be interpreted as follows:
\begin{equation}
\begin{aligned}
\label{eq:v3}
\mathcal{V}_i^{\mathcal{E},b} =\{p \in (\mathcal{E} \setminus \mathcal{O}) |\quad & a_{ij}^{\mbox{\footnotesize T}} p\leq b_{ij}-\beta_{ij},\forall j\neq i, i,j\in \mathcal{I},\\
 \quad & a_{io}^{\mbox{\footnotesize T}} p\leq b_{io}-\beta_{io},o\in \mathcal{I}_o\\
\quad & a_{ie}^{\mbox{\footnotesize T}} p\leq b_{ie} \}. \\
\mathcal{V}_e^{\mathcal{E},b} =\{p \in (\mathcal{E} \setminus \mathcal{O}) |\quad & a_{eo}^{\mbox{\footnotesize T}} p\leq b_{eo},\\
\quad & a_{ei}^{\mbox{\footnotesize T}} p\leq b_{ei} \}.
\end{aligned} 
\end{equation}





It can be observed that if a pursuer is in a collision-free configuration, i.e., for initialization $d_{ij}(0) \geq r_i+r_j, \forall i \neq j$, $d_{io}(0) \textgreater r_i$, then the OA-ECBVC can be generated.
If we design a control strategy such that the positions of the pursuer satisfy the constraints in Eq.~(\ref{ai}) and Eq.~(\ref{ao}) for all future time, then we can conclude that the planned pursuit path will remain collision-free, with $d_{ij}(t) \geq r_i + r_j$ for all $i \neq j$ and $0 \leq t \leq t_c$, and $d_{io} > r_i$ at all times.

\section{Encirclement and Capture Strategy}
\label{sec:ec}
\subsection{Encirclement Strategy}

A cooperative encirclement strategy based on the OA-ECBVC $\mathcal{V}_i^{\mathcal{E},b}$ is required for each pursuer $i$ to scatter around the evader and block its movements. 
The performance of encirclement by a team of pursuers $p_1,...,p_n$ is assessed using an energy function $\mathcal{H}(\mathcal{E},\mathcal{P})$ with constant density \cite{ref37}:
\begin{equation}
\begin{aligned}
\label{energy}
\mathcal{H}(\mathcal{P}) = \sum_{i=1}^{n} \mathcal{H}_i(\mathcal{P})
=\sum_{i=1}^{n}\int_{\mathcal{V}_i^{\mathcal{E},b}}\Vert p-p_i \Vert_{2}^2 dp.
\end{aligned} 
\end{equation}
 The partial derivative of $\mathcal{H}(\mathcal{P})$ with respect to the position of the $i$ pursuer is that
\begin{equation}
\begin{aligned}
\label{mc}
\frac{\partial \mathcal{H}(\mathcal{P})}{\partial p_i} 
= M_{\mathcal{V}_i^{\mathcal{E},b}}( p_i-C_{\mathcal{V}_i^{\mathcal{E},b}}).
\end{aligned} 
\end{equation}
 where $M_{\mathcal{V}_i^{\mathcal{E},b}}$ and $C_{\mathcal{V}_i^{\mathcal{E},b}}$ are the mass and mass centroid of each OA-ECBVC. We can see that the value of $\frac{\partial \mathcal{H}(\mathcal{P})}{\partial p_i} $ is only determined by the position of the pursuer and its Voronoi neighbors. The most favorable locations for the pursuers, which minimize the value of $\mathcal{H}$, are situated at the centroid of their corresponding Voronoi cells.

When pursuing the evader, each pursuer needs to update $\mathcal{V}_i^{\mathcal{E},b}$ according to $p_e$ and $C_{\mathcal{V}_i^{\mathcal{E},b}}$ will be changed with time $t$ correspondingly.
The encirclement strategy should drive each pursuer's location converge to its mass centroid of the Voronoi cell, such that $p_i=C_{\mathcal{V}_i^{\mathcal{E},b}}$. In the following section, we will introduce the detailed design process using various dynamic models to guide pursuers toward their centroid.

If the pursuers have greater dynamic limits than the evader, the movement of the pursuer's centroid will be less than the distance traveled by the evader,
such that $p_i$ will converge to  $C_{\mathcal{V}_i^{\mathcal{E},b}}$ eventually according to  clustering theory. 
As a result, a Centroidal Voronoi Tessellation (CVT) \cite{ref29} is constructed where all generators $p_e,p_1,\dots,p_n$ coincide with their $C_{\mathcal{V}_e^{\mathcal{E},b}},C_{\mathcal{V}_1^{\mathcal{E},b}},\dots,C_{\mathcal{V}_n^{\mathcal{E},b}}$. 
 The positions of all pursuers are evenly distributed around the center \cite{song2013distributed}.
 
 According to the statement about standard Voronoi partition $\mathcal{V}(\mathcal{P})$ in \cite{ref42}, 	$p_i$ is a vertex of the convex hull of the set $\mathcal{P}$ if and only if its corresponding  Voronoi cell shares half-infinite rays (edges) with its neighboring Voronoi cells. 
Since the $\mathcal{V}_e$ is located at the center of $\mathcal{E}$, there are $n_k$ Voronoi cells with $n_k \leq n$ sharing semi-infinite rays with their neighbors in $\mathcal{V}(\mathcal{P})$. Therefore, the pursuers that correspond to these $k$  Voronoi cells, i.e., $z_1,\dots,z_{8}$, 
can form a convex hull. 


\subsection{Capture Strategy}

Apart from encircling the evader, the pursuers are also required to decrease the distances between them and the evader rapidly.
Therefore, the capture strategy is designed to decrease $d_c$ by continually shrinking the boundary of $\mathcal{E}$ once $p_e \in \Omega$. 
The shrinkage amount, defined as $D(\Delta t)$, is determined by the movement distance of pursuer $i$ during a time step $\Delta t$. 
Then,
$\Delta x(t)$ and $\Delta y(t)$ during $t_e \leq t\leq t_c$ can be updated using the following recursive equations:
\begin{equation}
\begin{aligned}
\label{eq: captureElength}
& \Delta x(t+\Delta t) =\Delta x(t)- D(\Delta t), \\
& \Delta y(t+\Delta t) =\Delta y(t)- D(\Delta t).
\end{aligned} 
\end{equation}
Shrinking $\mathcal{E}$ will cause the centroids of Voronoi cells to shift inwards the region. 
When pursuers move to the centroids of their OA-ECBVC, the area of each cell in $\mathcal{V}(\mathcal{P},\mathcal{E},\mathcal{O})$ will tend to become even according to above discussion. Therefore, as the entire area of $\mathcal{E}$ shrinks with time, the Voronoi area of pursuers will continually become more uniform and smaller. It will further decrease the area of $\mathcal{V}^{\mathcal{E},b}_e$, due to the evader being encircled by pursuers. 
Therefore, $d_c$ will also be reduced until satisfying $d_c \leq r_c$. 



\section{Cooperative Pursuit using OA-ECBVC with Collision Avoidance}
\label{sec:alg}
Our cooperative pursuit strategy with collision avoidance method using the OA-ECBVC is outlined in this section. We utilize reactive feedback control for the single integrator and model predictive control for the triple  integrator dynamic of pursuers to  guide pursuers toward their centroid.

\subsection{Single-integrator Dynamics}
Consider a pursuer using a dynamic model with a single integrator:
\begin{equation}
\label{model}
\begin{aligned}
\dot{p}_i=v_i, &\quad \Vert v_i \Vert_{2} \leq v_{p,\max}.
\end{aligned}
\end{equation}
where $v_i$ is velocity control inputs of the pursuer subjected to its maximum speed $v_{p,\max}$.
 We can use a dissipative reactive control law $v_i$, in which each pursuer $i$ follows its negative gradient component and moves over its dominance region $\mathcal{V}_i^{\mathcal{E},b}$ \cite{ref37}.
\begin{equation}
\begin{aligned}
\label{u}
v_i = -\Vert v_{p,\max} \Vert_{2} \frac{p_i - C_{\mathcal{V}_i^{\mathcal{E},b}}}{\Vert p_i - C_{\mathcal{V}_i^{\mathcal{E},b}} \Vert_{2}},
\end{aligned} 
\end{equation}
At each time step, each pursuer constructs its OA-ECBVC using Eq.~(\ref{eq:v3}) and computes its centroid. Note that the constructed OA-ECBVC is a convex polytope. Therefore, finding the centroid of a polytope can be efficiently solved using linear algebra, which has a time complexity linearly related to the number of vertices.

\subsection{Triple-integrator Dynamics}
\label{sec:tri}
Consider a pursuer using a dynamic model with a triple integrator:
\begin{equation}
\begin{aligned}
\dot{p_i}=v_i, &\quad \Vert v_i \Vert_{2} \leq v_{p,\max}\\
\dot{v_i}=a_i, &\quad \Vert a_i \Vert_{2} \leq a_{p,\max}\\
\dot{a_i}=j_i, &\quad \Vert j_i \Vert_{2} \leq j_{p,\max}
\end{aligned}
\end{equation}
where $a_i$, $j_i$ are its acceleration and jerk subjected to the maximum value $a_{p,\max}$ and $j_{p,\max}$ respectively. $\mathrm{x}_i=[p_i, v_i, a_i]^{\mbox{\footnotesize T}}$ denotes the state of the robot.



To create a trajectory that passes through the centroid continuously, a local trajectory generation technique is utilized with the help of motion primitives $\zeta(\mathrm{x}_i^s,\mathrm{x}_i^g,u_i)$  \cite{Shupeng}, which are capable of steering the system from a starting state $\mathrm{x}_i^s$ to a goal state $\mathrm{x}_i^g$, where $u_i$ is the control input.
A nonlinear optimization problem can be formulated based on an model predictive control (MPC) method to select the best motion primitive with $t\in[t_0,t_0+\Delta t]$.

 \begin{equation}
\begin{aligned}
\label{eq: opt}
\min \quad &\int_{t_0}^{t_0+\Delta t} \lambda_u J_{u} + \lambda_p J_{p} \quad dt \\
\mbox{s.t.}\quad & \mathrm{x}_i(t_0) = \mathrm{x}_i^s, \mathrm{x}_i(t_0+\Delta t) = \mathrm{x}_i^g\text{,}\\
& \dot{p_i}=v_i, \dot{v_i}=a_i, \dot{a_i}=j_i\text{,}\\
& \mathrm{x}_i \in \mathcal{V}_i^{\mathcal{E},b}\text{,}\\
& u_i \in \mathcal{U}_i\text{.}
\end{aligned}
\!\!\!\!\!\!\!\!\!\!
\end{equation}
where $\mathcal{U}_i$ is admissible control space. 
$J_{u}$ and $J_{f}$ denote the flight smoothness and goal-reaching performance, respectively, where $\lambda_{u}$, $\lambda_{f}$ are the trade-off terms. 1) $J_{u} = ||u_i||_2^2$ penalize the aggressive trajectories.
2) $J_{p}=\Vert \zeta(\mathrm{x}_i^s,\mathrm{x}_{i}^g,u_i)-C_{\mathcal{V}_i^{\mathcal{E},b}}\Vert_2^2 $ represents the performance of navigating to centroid points. The cost function will explicitly penalize the contouring error between the trajectory that needs to be optimized and desired centroid point.
By solving this problem using our previous method in \cite{xilele}, a dynamically feasible trajectory can be quickly generated to guide the pursuer following along the pursuit path in a receding horizon fashion. 


\subsection{Method Overview}


\begin{figure}[!t]
	\centering
	\setlength{\belowcaptionskip}{-0.2cm}
	\includegraphics[width=\hsize]{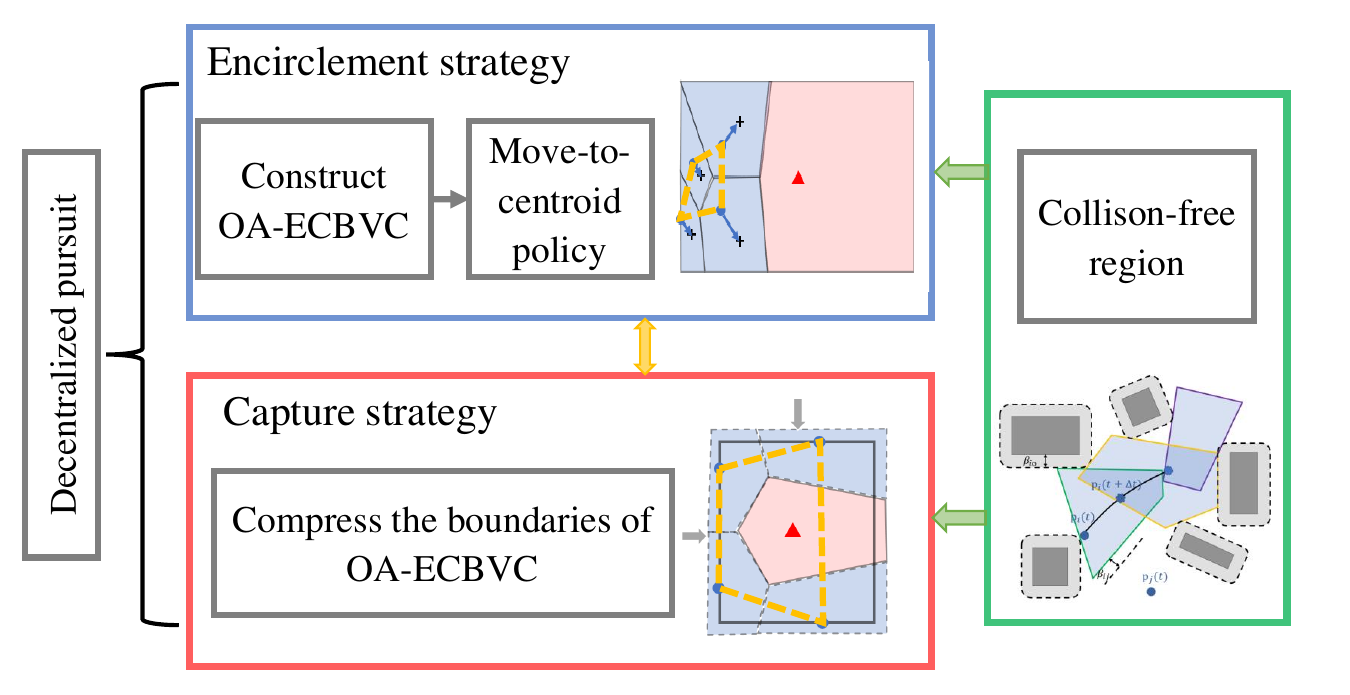}
	\caption{An overview of our algorithm.}
	\vspace{-8pt}
	\label{alg}
\end{figure}

Our algorithm relies on constantly constructing the OA-ECBVC to calculate pursuit policy in a cooperative way. The overall structure is shown in Fig.~\ref{alg}. In each replanning process, under the encirclement strategy, the team of pursuers tend to block the evader while approaching it. 
Each pursuer calculates the centroid of OA-ECBVC $C_{\mathcal{V}_i^{\mathcal{E},b}}$ and updates its position via Eq.~(\ref{u}) or solving the trajectory generation in Eq.~(\ref{eq: opt}) for the next moment.
Once the evader is encircled in a convex hull $\Omega$ formed by a team of pursuers, under the capture strategy, the boundaries of $\mathcal{E}$ will be compressed via Eq.~(\ref{eq: captureElength}) to rapidly decrease the evader's movement space. 	

\section{Simulations}
\label{sec: sim}
This section provides several trials with randomized initial configurations and evaluation metrics to measure the performance of collision avoidance through obstacle-rich areas.
Comparisons are then conducted with other state-of-the-art methods in terms of capture and encirclement efficiency. 

All numerical simulations are conducted in MATLAB environment under the same settings: 
$n$ pursuers capture one evader in an unbounded environment with cluttered obstacles. The safety radius for pursuer and evader is set as $r_i = 0.15$m, $r_e = 0.15$m, respectively. The capture radius is set as $r_c = 1.0$m. 
The replanning time step $\Delta t = 0.1$s.
The maximum velocity for all pursuers and the evader is set as $v_{p,\max} = 1$m/s and $v_{e,\max} = 0.9$m/s, respectively.

To assess  the applicability of our method, we choose the escape policy for the evader described  in \cite{ref15}, that the evader moves to the centroid of standard Voronoi cell $C_{\mathcal{V}_e^{\mathcal{W}}}$ in the environment $\mathcal{W}$. This policy makes the evader far away from the neighbor pursuers by moving to its Voronoi centroid, which is designed as follows:  
\begin{equation}
\begin{aligned}
\label{MCP}
u_{e} = \Vert v_{e,\max} \Vert_{2} \frac{C_{\mathcal{V}_e^{\mathcal{W}}}-p_e}{\Vert C_{\mathcal{V}_e^{ \mathcal{W}}}-p_e \Vert_{2}}.
\end{aligned} 
\end{equation}
 For higher-order dynamics, we use the same technique described in Section~\ref{sec:tri} to follow the generated escaping path, and the maximum acceleration and jerk are set to be the same for both pursuers and evader.
\subsection{Performance Evaluation}
\label{eval}

\begin{figure}[!t]
\centering
\setlength{\belowcaptionskip}{-0.5cm}
\setlength{\abovecaptionskip}{0.4cm}		\hspace{-0.4cm}
\vspace{-3pt}
\subfloat[Initial time $t$=0.0s]{
\includegraphics[width=1.79in]{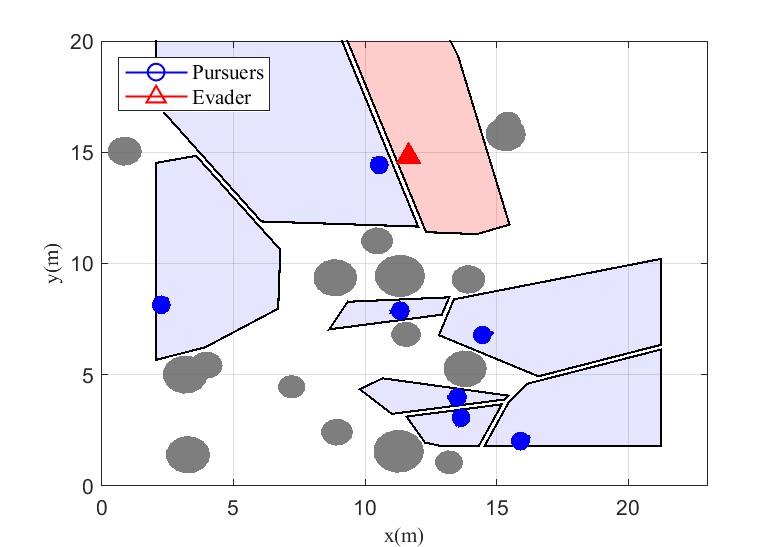}}
\hfil
\hspace{-0.71cm}
\vspace{-3pt}
\subfloat[Capture time $t_c$=15.0s]{
\includegraphics[width=1.79in]{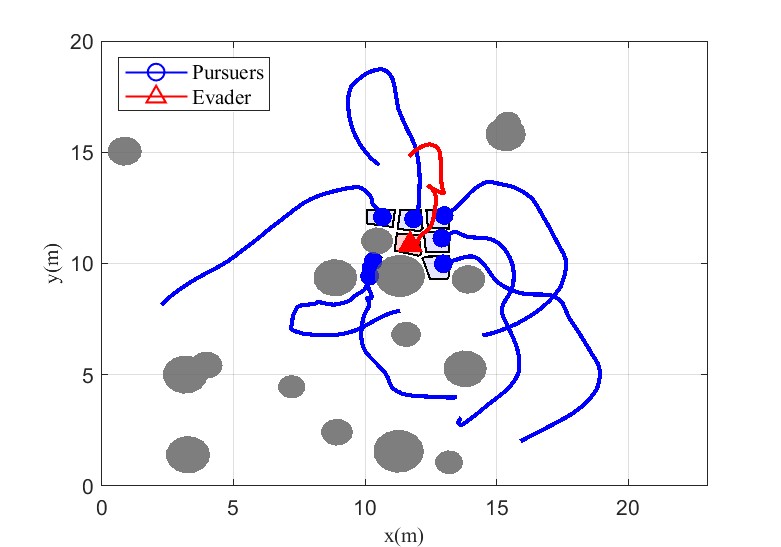}}
\hfil
\caption{Trajectories of 7 pursuers successfully capture one evader in forest environments with 16 obstacles for a non-encircled initial configuration. The OA-ECBVC of each pursuer and the evader are shaded in blue and red, respectively, shrinking as time passes.}
\label{fig:perf}
\end{figure}

We validate the effectiveness of our pursuit method in a large-scale unbounded environment that includes randomly generated obstacles to simulate real-world forest environments, as depicted in Fig.~\ref{fig:perf}.
The encirclement strategy employed by the pursuers involves calculating a smooth trajectory while moving in different directions to restrict the evader's movement. Once the evader is blocked, the pursuers resize the boundaries of their OA-ECBVC to rapidly reduce $d_c$ until capture it.
The dense obstacles within the forest pose a significant challenge to the pursuers but thanks to our collision-free Voronoi cells and MPC-based motion planner, pursuers can smoothly navigate through the forest while  effectively pursuing the evader. 
 Although the escape policy makes the evader react responsively to pursuers in terms of the threat posed by a whole team, the evader can still be successfully caught, as shown in Fig.~\ref{fig:perf}(b). 
 
\begin{figure}[!t]
\centering
\setlength{\belowcaptionskip}{-0.5cm}
\setlength{\abovecaptionskip}{0.4cm}		\hspace{-0.4cm}
\vspace{-3pt}
\subfloat[Collision distance over time]{
\includegraphics[width=1.79in]{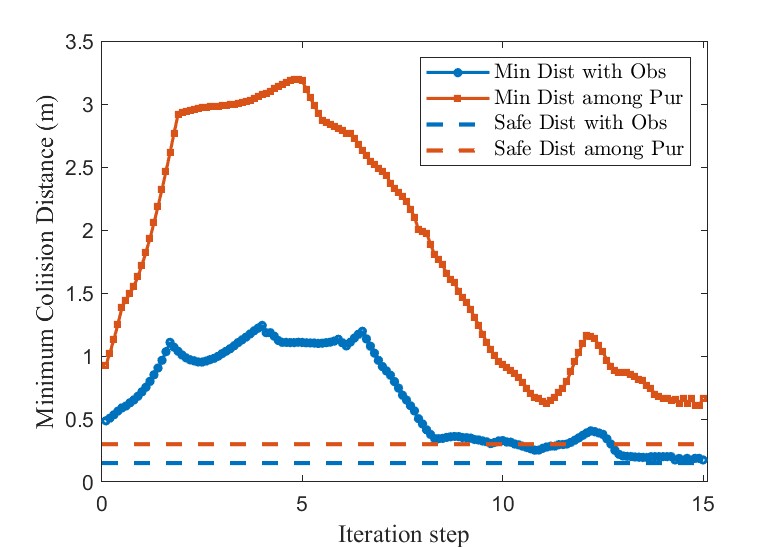}}
\hfil
\hspace{-0.71cm}
\vspace{-3pt}
\subfloat[Minimum collision distance]{
\includegraphics[width=1.79in]{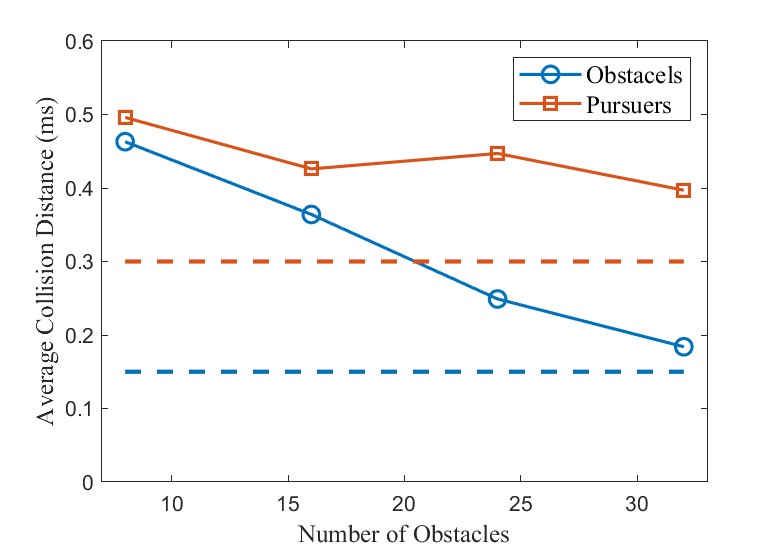}}
\hfil
\caption{Evaluation of collision distance with obstacle and a whole team.}
\label{fig_collision}
\end{figure}

 In addition, we use the minimum collision distance among a team of pursuers and obstacles in Eq.~(\ref{distobs}) $d_{mo} 
 = \min_{i\in \mathcal{I}} d_{io}$ and Eq.~(\ref{distpur}) $d_{mp} = \min_{i,j\in \mathcal{I}} d_{ij}$ to show safety ratio of the above trajectory over time, as illustrates in Fig.~\ref{fig_collision}(a). All pursuers can always remain within their respective safety areas in the OA-ECBVC during pursuit ($d_{mo}$=0.17m and $d_{mp}$=0.61m). We further measure the mean value of $d_{mo}$ and $d_{mp}$ by executing 40 trials with random obstacles and initial configurations for all robots. As shown in Fig.~\ref{fig_collision}(b), even for a large number of obstacles, the pursuers still maintain a safe distance from both each other and the obstacles, ensuring a high safety ratio. 
 

\subsection{Comparative Results}

To validate the performance in terms of encirclement and capture efficiency,  we make a comparison with the other two algorithms:
1) Obstacle-aware Voronoi cell (OAVC) method proposed in \cite{ref15}, in which pursuers chase an evader guided by an information density map of the evader.
2) Surrounding and hunting (SH) method proposed in \cite{ref26}, which designs a distributed control law for  surrounding and hunting. In both OAVC and SH methods, the robot is considered a single-integral dynamic system, therefore we keep the same configuration in our comparison.

\begin{figure}[!t]
\centering
\setlength{\belowcaptionskip}{-0.5cm}
\setlength{\abovecaptionskip}{0.4cm}
\hspace{-0.45cm}
\vspace{-3pt}
\subfloat[Our method with $t_c=24.40$s]{
\includegraphics[width=1.75in]{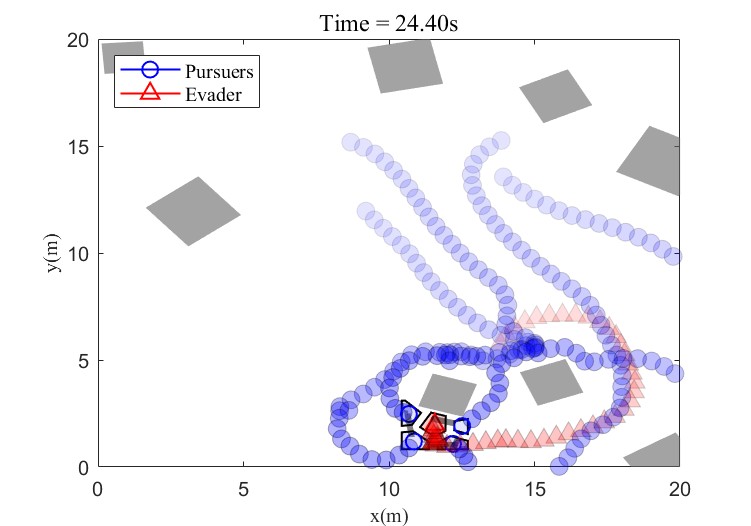}
\label{oavc1}}
\hfil
\hspace{-0.8cm}
\vspace{-3pt}
\subfloat[OAVC method with $t_c=32.20$s]{
\includegraphics[width=1.75in]{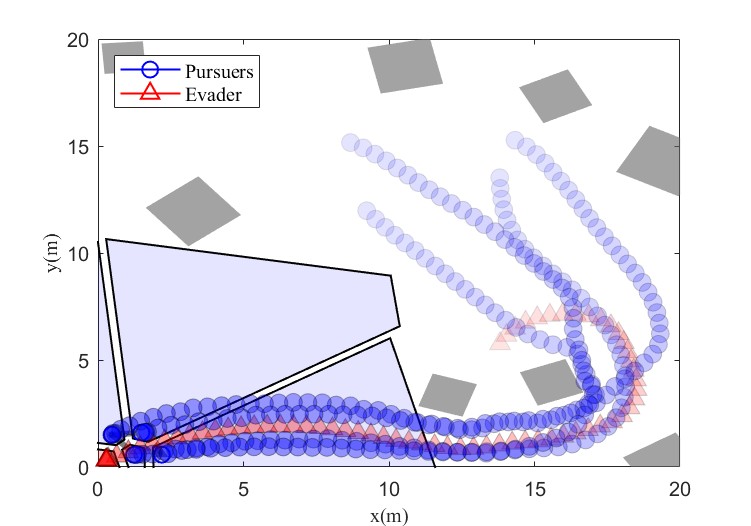}
\label{oavc2}}
\hfil

\hspace{-0.45cm}
\vspace{-3pt}
\subfloat[Our method with $t_c = 15.10$s]{
\includegraphics[width=1.75in]{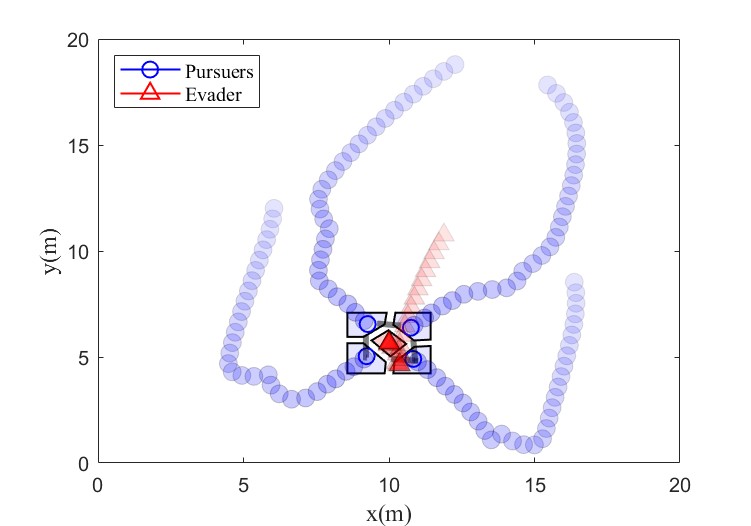}
\label{sh1}}
\hfil
\hspace{-0.8cm}
\vspace{-3pt}
\subfloat[SH method with $t_c = 18.20$s]{
\includegraphics[width=1.75in]{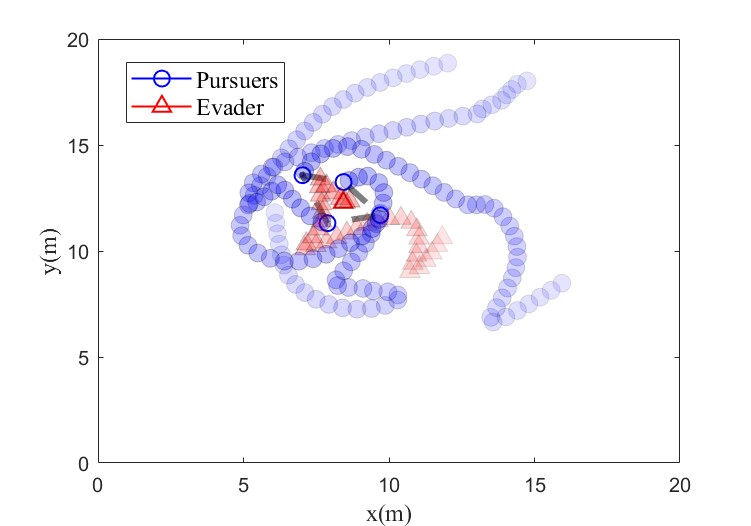}
\label{sh2}}
\hfil
\caption{Pursuit path obtained by our method, OAVC method, and SH method. The grey dotted line represents the encirclement formed by a team of pursuers. 
(a)-(b) OAVC fails to encircle the evader, leading to a longer capture time. (c)-(d) SH has lower capture efficiency although it can maintain encirclement in an obstacle-free environment.
}
\label{1compare}
\end{figure}

As OAVC cannot handle convex polygon obstacles, we modified its Voronoi construction method so that it can be applied to the comparison scenarios.
As shown in Fig.~\ref{1compare}(b), OAVC method failed to trap the evader before capturing it.
This is because the OAVC method drives the pursuers to move toward the high-density value of the evader in an unbounded environment, which is somewhat equivalent to moving directly toward the evader. This strategy may be too greedy and inadvertently create opportunities for the evader to escape. In contrast, our encirclement strategy allows the pursuers to react more intelligently to the evader, resulting in more efficient capture. 

\begin{figure}[!t]
	\centering
	\setlength{\belowcaptionskip}{-0.4cm}
	\includegraphics[width=2.5in]{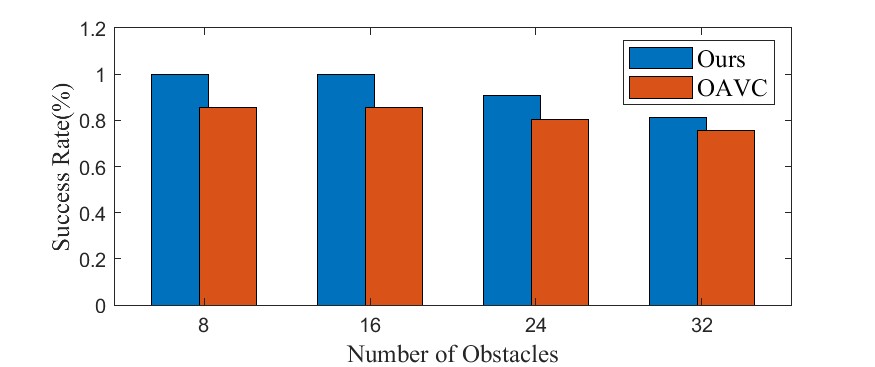}
	\caption{Comparisons of success rate with increasing the number of obstacles.}
	\vspace{-8pt}
	\label{success}
\end{figure}

We conducted additional experiments to evaluate the success rate in capturing evaders within a given range (19.8m for both the x-axis and y-axis), comparing it to the OAVC method with increasing numbers of obstacles, as shown in Fig.~\ref{success}.
40 initial configurations for four pursuers randomly generated within region $[10,15] \times [10,15]$ in an unbounded and cluttered environment (20m$\times$20m). 
Using the OAVC method, pursuers targeting the evader must circumvent obstacles to apprehend them. The complexity of these barriers often provides the evader with numerous escape possibilities. Conversely, our algorithm is more effective in preventing evaders from escaping, resulting in higher success rates.


SH method needs to spend more time  blocking the evader, as shown in Fig.~\ref{1compare}(d). This is because the efficiency and success rate of the surrounding is related to their relative initial positions.
In contrast, our method can always keep the evader in the encirclement condition using the control law in Eq.~(\ref{u}) with shorter $t_c$ than SH method.

\section{Experiments}
\label{sec6}
In this section, we demonstrate the effectiveness and robustness of the proposed algorithm in a real-world environment since mechanics and motor properties, localization uncertainty, and frictions cannot be accurately modeled in simulations and will influence the behavior of the pursuit.

In our decentralized multi-robot systems, each robot 
is equipped with an embedded computing platform (Intel i7 CPU@2.60 GHz) running ROS, allowing the robot to calculate its policy autonomously in real time. 
Information was transferred through the ROS network, and each pursuer only need to obtain the position of other robots via an external motion capture system (NOKOV) without knowing their policies. 
The speed $v_{p,\max} = 0.3$m/s for each pursuer and $v_{e,\max}= 0.2$m/s for the evader. The $r_c$ is set to be 1m with robot radius $r_i = 0.3$m, $i\in \mathcal{I}$, and $r_e = 0.3$m. The replanning time step $\Delta t$ is 0.25s. The full video can be found at https://youtu.be/wDxYuEJ1cKs.

\begin{figure}[!t]
	\centering	
	\hspace{-0.1cm}
	\vspace{-1.5pt}
	\subfloat[$t$=0.0s]{\includegraphics[width=1.43in]{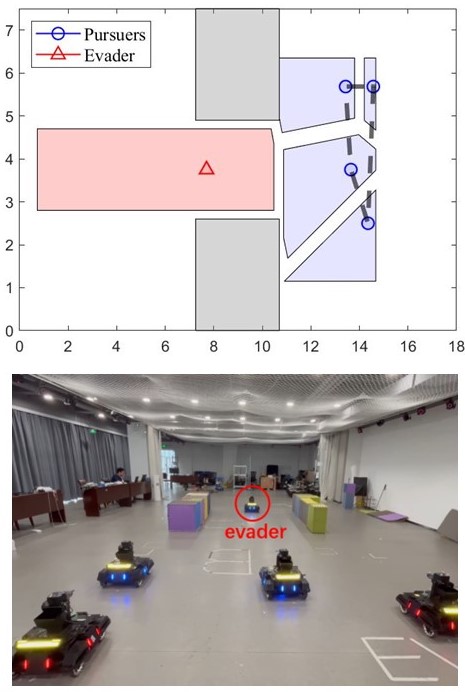}}
	\hfil
	\hspace{0.1cm}
	\vspace{-1.5pt}
	\subfloat[$t_c$=24.02s]{
		\includegraphics[width=1.45in]{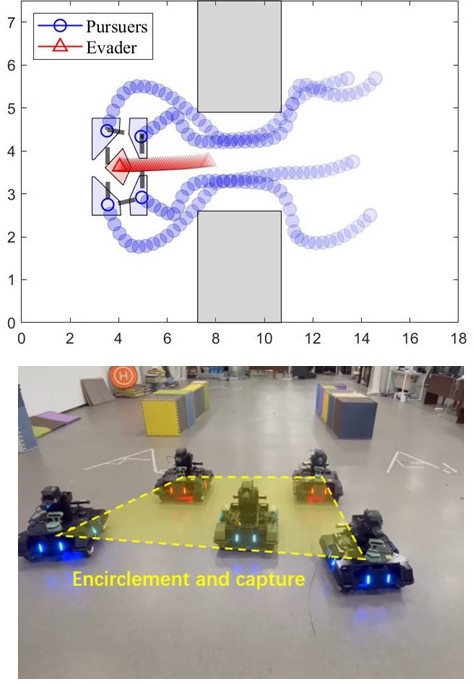}
		}
	\hfil
		\hspace{-0.1cm}
	\caption{Four pursuers pursue an autonomous evader in a narrow space environment with the size of 8m$\times$16m.} 
	\label{expadv2}
\end{figure}

We choose a tunnel scenario due to the limited space can easily cause multiple robots to get stuck or shake back and forth if they cannot cooperate in finding directions.
Additionally, the potential presence of obstacles and hazards within the tunnel can add an extra layer of difficulty to the pursuit.
As shown in Fig.~\ref{expadv2}(a), a group of pursuers is initially at one side of the corridor.  Although the pursuers are positioned together at first, they try to spread out around the evader while approaching it. In Fig.~\ref{expadv2}(b), they smoothly navigate through the corridor and cooperatively approach the evader by our method.  Once coming out of the corridor, pursuers adaptively adjust the strategy to encircle the evader. Eventually, the evader is captured with $t_c=24.02$s after the encirclement.
Despite the presence of noisy actuation and localization uncertainties, a team of pursuers can still safely navigate through narrow spaces and make an efficient strategy, which demonstrates the reliability of our method in terms of real-time performance.





\section{Conclusions}
\label{sec:con}
This paper develops a decentralized Voronoi-based encirclement and capture algorithm for pursuing an evader in an unbounded environment with dense obstacles. 
Comparisons show that our algorithm can achieve a shorter capture time of around 20 $\%$ than other state-of-art methods in both obstacles and empty environments and has a higher success rate. 
Pursuing in complex environments with guaranteed collision avoidance also demonstrates the effectiveness and robustness of our method. Meanwhile, our method maintains a high safety ratio even when facing dense obstacles. The limitation lies in the pursuer's requirement of global position information of other players for coordination.
Future work includes extensions to multiple evaders with various unknown escape policies and local information.



\bibliographystyle{IEEEtran}
\bibliography{refs}

\begin{thebibliography}{10}
\providecommand{\url}[1]{#1}
\csname url@samestyle\endcsname
\providecommand{\newblock}{\relax}
\providecommand{\bibinfo}[2]{#2}
\providecommand{\BIBentrySTDinterwordspacing}{\spaceskip=0pt\relax}
\providecommand{\BIBentryALTinterwordstretchfactor}{4}
\providecommand{\BIBentryALTinterwordspacing}{\spaceskip=\fontdimen2\font plus
\BIBentryALTinterwordstretchfactor\fontdimen3\font minus \fontdimen4\font\relax}
\providecommand{\BIBforeignlanguage}[2]{{%
\expandafter\ifx\csname l@#1\endcsname\relax
\typeout{** WARNING: IEEEtran.bst: No hyphenation pattern has been}%
\typeout{** loaded for the language `#1'. Using the pattern for}%
\typeout{** the default language instead.}%
\else
\language=\csname l@#1\endcsname
\fi
#2}}
\providecommand{\BIBdecl}{\relax}
\BIBdecl

\bibitem{ref0}
T.~H. Chung, G.~A. Hollinger, and V.~Isler, ``Search and pursuit-evasion in mobile robotics,'' \emph{Autonomous Robots}, vol.~31, no.~4, pp. 299--316, 2011.

\bibitem{semnani2017multi}
S.~H. Semnani and O.~A. Basir, ``Multi-target engagement in complex mobile surveillance sensor networks,'' \emph{Unmanned Systems}, vol.~5, no.~01, pp. 31--43, 2017.

\bibitem{ref1}
C.~Robin and S.~Lacroix, ``Multi-robot target detection and tracking: taxonomy and survey,'' \emph{Autonomous Robots}, vol.~40, no.~4, pp. 729--760, 2016.

\bibitem{khan2022novel}
S.~Khan, M.~Tufail, M.~T. Khan, Z.~A. Khan, J.~Iqbal, and A.~Wasim, ``A novel framework for multiple ground target detection, recognition and inspection in precision agriculture applications using a {UAV},'' \emph{Unmanned Systems}, vol.~10, no.~01, pp. 45--56, 2022.

\bibitem{ref2}
A.~Khan, B.~Rinner, and A.~Cavallaro, ``Cooperative robots to observe moving targets,'' \emph{IEEE Transactions on Cybernetics}, vol.~48, no.~1, pp. 187--198, 2016.

\bibitem{ref3}
R.~Isaacs, \emph{Differential games}.\hskip 1em plus 0.5em minus 0.4em\relax New York: Wiley, 1967.

\bibitem{ref4}
T.~Ba{\c{s}}ar and G.~J. Olsder, \emph{Dynamic noncooperative game theory}.\hskip 1em plus 0.5em minus 0.4em\relax London: SIAM, 1998.

\bibitem{ref5}
H.~Huang, J.~Ding, W.~Zhang, and C.~J. Tomlin, ``Automation-assisted capture-the-flag: a differential game approach,'' \emph{IEEE Transactions on Control Systems Technology}, vol.~23, no.~3, pp. 1014--1028, 2014.

\bibitem{ref15}
A.~Pierson, Z.~Wang, and M.~Schwager, ``Intercepting rogue robots: an algorithm for capturing multiple evaders with multiple pursuers,'' \emph{IEEE Robotics and Automation Letters}, vol.~2, no.~2, pp. 530--537, 2016.

\bibitem{ref16}
Z.~Zhou, W.~Zhang, J.~Ding, H.~Huang, D.~M. Stipanovi{\'c}, and C.~J. Tomlin, ``Cooperative pursuit with voronoi partitions,'' \emph{Automatica}, vol.~72, pp. 64--72, 2016.

\bibitem{ref17}
H.~Huang, W.~Zhang, J.~Ding, D.~M. Stipanovi{\'c}, and C.~J. Tomlin, ``Guaranteed decentralized pursuit-evasion in the plane with multiple pursuers,'' in \emph{Proceedings of 50th IEEE Conference on Decision and Control and European Control Conference}, 2011, pp. 4835--4840.

\bibitem{pierson2017distributed}
A.~Pierson and D.~Rus, ``Distributed target tracking in cluttered environments with guaranteed collision avoidance,'' in \emph{2017 International Symposium on Multi-Robot and Multi-Agent Systems (MRS)}.\hskip 1em plus 0.5em minus 0.4em\relax IEEE, 2017, pp. 83--89.

\bibitem{ref20}
B.~Tian, P.~Li, H.~Lu, Q.~Zong, and L.~He, ``Distributed pursuit of an evader with collision and obstacle avoidance,'' \emph{IEEE Transactions on Cybernetics}, 2021, early Access.

\bibitem{ref21}
D.~Zhou, Z.~Wang, S.~Bandyopadhyay, and M.~Schwager, ``Fast, on-line collision avoidance for dynamic vehicles using buffered voronoi cells,'' \emph{IEEE Robotics and Automation Letters}, vol.~2, no.~2, pp. 1047--1054, 2017.

\bibitem{ref25}
C.~Wang, H.~Chen, J.~Pan, and W.~Zhang, ``Encirclement guaranteed cooperative pursuit with robust model predictive control,'' in \emph{Proceedings of IEEE/RSJ International Conference on Intelligent Robots and Systems}, 2021, pp. 1473--1479.

\bibitem{ref26}
X.~Fang, C.~Wang, L.~Xie, and J.~Chen, ``Cooperative pursuit with multi-pursuer and one faster free-moving evader,'' \emph{IEEE Transactions on Cybernetics}, vol.~52, no.~3, pp. 1405--1414, 2022.

\bibitem{ref35}
Z.~Zhang, X.~Wang, Q.~Zhang, and T.~Hu, ``Multi-robot cooperative pursuit via potential field-enhanced reinforcement learning,'' \emph{arXiv preprint arXiv:2203.04700}, 2022.

\bibitem{ref33}
J.~Liao, C.~Liu, and H.~H. Liu, ``Model predictive control for cooperative hunting in obstacle rich and dynamic environments,'' in \emph{Proceedings of IEEE International Conference on Robotics and Automation}, 2021, pp. 5089--5095.

\bibitem{ref27}
O.~Arslan and D.~E. Koditschek, ``Sensor-based reactive navigation in unknown convex sphere worlds,'' \emph{The International Journal of Robotics Research}, vol.~38, no. 2-3, pp. 196--223, 2019.

\bibitem{ref37}
Q.~Du, V.~Faber, and M.~Gunzburger, ``Centroidal voronoi tessellations: Applications and algorithms,'' \emph{SIAM Review}, vol.~41, no.~4, pp. 637--676, 1999.

\bibitem{ref29}
J.~Cortes, S.~Martinez, T.~Karatas, and F.~Bullo, ``Coverage control for mobile sensing networks,'' \emph{IEEE Transactions on Robotics and Automation}, vol.~20, no.~2, pp. 243--255, 2004.

\bibitem{song2013distributed}
Y.~Song, B.~Wang, Z.~Shi, K.~R. Pattipati, and S.~Gupta, ``Distributed algorithms for energy-efficient even self-deployment in mobile sensor networks,'' \emph{IEEE Transactions on Mobile Computing}, vol.~13, no.~5, pp. 1035--1047, 2013.

\bibitem{ref42}
F.~P. Preparata and M.~I. Shamos, \emph{Computational geometry: an introduction}.\hskip 1em plus 0.5em minus 0.4em\relax New York: Springer Science \& Business Media, 2012.

\bibitem{Shupeng}
S.~Lai, M.~Lan, and B.~M. Chen, ``Model predictive local motion planning with boundary state constrained primitives,'' \emph{IEEE Robotics and Automation Letters}, vol.~4, no.~4, pp. 3577--3584, 2019.

\bibitem{xilele}
L.~Xi, X.~Wang, L.~Jiao, S.~Lai, Z.~Peng, and B.~M. Chen, ``{GTO-MPC}-based target vhasing using a quadrotor in cluttered environments,'' \emph{IEEE Transactions on Industrial Electronics}, vol.~69, no.~6, pp. 6026--6035, 2022.

\end{thebibliography}

\end{document}